\documentclass[journal]{IEEEtran}

\usepackage{amsmath,amsfonts}
\usepackage{algorithmic}
\usepackage{algorithm}
\usepackage{array}
\usepackage[caption=false,font=normalsize,labelfont=sf,textfont=sf]{subfig}
\usepackage{textcomp}
\usepackage{stfloats}
\usepackage{url}
\usepackage{verbatim}
\usepackage{graphicx}
\usepackage{cite}
\usepackage[inline]{enumitem} 
\usepackage{tikz} 
\usepackage{pgfplots} 
\pgfplotsset{compat=1.18}
\usepackage{booktabs}
\usetikzlibrary{3d,calc} 
\usepackage[dvipsnames]{xcolor} 

\usepackage[final]{microtype}
\DeclareMathOperator*{\argmax}{arg\,max}
\DeclareMathOperator*{\cosim}{cosim}

\begin{document}
	
    \title{Contrastive Learning for Seismic Horizon Tracking with Domain-Specific Priors
    }

    \author{Alexandre Thouvenot,
            Lionel Boillot,
            and Vincent Gripon,~\IEEEmembership{Senior Member, IEEE}%
    \thanks{Alexandre Thouvenot and Vincent Gripon are with IMT Atlantique, LAB-STICC,
    UMR CNRS 6285, F-29238 Brest, France (email: alexandre.thouvenot@imt-atlantique.fr; vincent.gripon@imt-atlantique.fr).}%
    \thanks{Alexandre Thouvenot and Lionel Boillot are with TotalEnergies, OneTech,
    64000 Pau, France (email: alexandre.thouvenot@totalenergies.com; lionel.boillot@totalenergies.com).}%
    }

    \markboth{}%
    {Thouvenot \textit{et al.}: Contrastive Learning for Seismic Horizon Tracking with Domain-Specific Priors}

    \maketitle
	
    \begin{abstract}

    Unsupervised 3D seismic horizon tracking faces a key limitation: signal-based propagators provide accurate trace-level alignment but often fail near faults, whereas texture-driven deep models are more robust to discontinuities, typically at the cost of labeled data requirements and reduced trace-level precision. We propose a self-supervised fusion of both paradigms in which signal-derived local horizon correspondences act as domain-specific priors to train a texture-based deep learning model. Specifically, we estimate reliable trace-to-trace flows from reflector slopes and use them to form positive pairs in a contrastive objective, while restricting training to high-confidence neighborhoods, optionally augmented with a fault mask. The objective is not to infer ambiguous correspondences close to discontinuities, but to preserve horizon identity across them. As a result, the network learns voxel-wise embeddings that preserve local signal continuity while enabling horizon propagation beyond discontinuities through similarity search. Experiments on the public F3 dataset and a faulted synthetic dataset achieve lower mean absolute error (MAE) than unsupervised baselines and competitive performance against a semi-supervised method using a single labeled slice.
    
    \end{abstract}
	
	
    \section{Introduction}
		
    Seismic data provide 3D images of subsurface geological structures. Geoscientists interpret various subsurface features such as faults, salt bodies, channels, and stratigraphic surfaces (hereafter referred to as horizons), which are the focus of this work. Accurately tracking horizons throughout a 3D seismic volume is critical for many applications, yet remains challenging, as their geometry is affected by folding, faulting, and sedimentary variations. Here, horizon texture refers to local temporal patterns along seismic traces, such as amplitude and frequency content, rather than spatial texture across neighboring traces. These patterns can vary significantly across the volume, creating a trade-off between trace-level precision and robustness to discontinuities. As a result, manual interpretation is time-consuming and requires expert knowledge.

    A wide range of methods has been developed to automate horizon interpretation, from signal-based propagation techniques~\cite{guillon2010Corr, Lomask2006rgt, abubakar2022rgtFlow, guillon2013geotime} to deep learning models~\cite{silva2019netherlands, Wang2023segFormer}. Signal-based approaches exploit local trace similarity (e.g., cross-correlation or amplitude-gradient methods) to propagate horizons, but they often fail near discontinuities because they assume local continuity. Deep learning approaches typically learn texture-based features and can be more robust for recovering horizons across faults or erosional breaks, yet they generally require labeled data and training sets that cover the full variability of horizon textures. This motivates methods that transfer the local precision of signal-based alignment into learned representations that can preserve horizon identity across structurally complex regions.

    In this work, we combine signal-based and texture-based approaches through a self-supervised framework with two components: \begin{enumerate*}[label=(\roman*)]
        \item an encoder that extracts voxel-wise features from seismic traces, and
        \item a contrastive objective that aligns embeddings along locally estimated trace-to-trace correspondences (signal-based) while separating different depths within a trace.
    \end{enumerate*}

    Our main contributions are:
    \begin{itemize}
        \item A self-supervised framework that learns voxel-wise embeddings for horizon tracking in 3D seismic volumes.
        \item A structure-informed contrastive pair definition: signal-based propagators define positive voxel pairs within high-confidence neighborhoods, while within-trace negatives enforce separation between horizons.
        \item Evaluations on a faulted synthetic dataset and the public F3 dataset, including ablations on architecture, embedding dimension, and faulted horizon propagation.
    \end{itemize}

    \section{Related Work}
    
    \subsection{Horizon Interpretation}
	
    \paragraph{Signal-Based Methods}
	
    The propagation of seismic horizons is a longstanding problem, often formulated as identifying local signal similarity between neighboring traces. Classical methods iteratively extend horizons from seed points by maximizing correlation~\cite{guillon2010Corr}. Although locally accurate, these approaches degrade in noisy or discontinuous regions.

    To mitigate correlation failures, alternatives such as Dynamic Time Warping~\cite{su2023dtw} and learning-based flow estimation for 1D trace alignment~\cite{li2021flowNet} (including 4D settings~\cite{dramsch2022voxelmorph4D}) have been explored. These methods can be more robust to certain discontinuities by explicitly modeling trace-to-trace deformation.
        
	Another widely used family relies on relative geological time (RGT) and slope estimation to follow seismic stratigraphy~\cite{Lomask2006rgt, abubakar2022rgtFlow, guillon2013geotime}. However, inaccurate slope estimates near faults may lead to poor horizon predictions in complex regions.
	
    \paragraph{Texture-Based Methods}
	
    Over the past decade, texture-based methods, generally implemented using deep learning architectures, have been increasingly explored~\cite{silva2019netherlands, Wang2023segFormer}. Their generalization remains limited by the representativeness of labeled training data.
    
	To reduce reliance on labeled data, semi-supervised approaches have gained attention. Prior work~\cite{thouvenot2025geo25d} introduced a 2D regularization constraint based on seismic slopes applied to unlabeled regions. While effective, its 2D formulation restricts the analysis to specific slices of the seismic cube, which may miss complex 3D structures. Other methods also explore contrastive learning to improve pixel-level predictions under limited supervision~\cite{Li2023conss}.
    
	More recently, pretrained seismic representation models have been transferred to downstream tasks such as horizon tracking~\cite{pham2025seisBERT}. While these features can improve propagation, performance may still depend on the diversity of the pretraining data and the extent to which the learned representations capture stratigraphic consistency.

    \subsection{Contrastive Learning}
	
    Contrastive learning with the InfoNCE objective~\cite{oord2019representationlearningcontrastivepredictive} has been extended to dense prediction, enabling pixel-level representation learning~\cite{xie2021pixelCl,goncharov2023voxel2vex}. These methods typically enforce consistency between corresponding pixels across different views as positive pairs, while treating unrelated pixels as negatives. In our setting, signal-based local propagation provides stratigraphically consistent correspondences that naturally define positive voxel pairs, while within-trace depth samples serve as negatives.
    
    \section{Methodology}

    We consider a 3D seismic volume of size $L \times W \times T$ represented as a set of traces $\mathbf{x}_{\mathbf{p}} \in \mathbb{R}^T$, indexed by spatial position $\mathbf{p} \in S = \{1, \dots, L\} \times \{1, \dots, W\}$. Stratigraphic horizons are modeled as functions $\nu_k : S \rightarrow \{1, \dots, T\}$ that assign a unique time sample to each spatial position, where $k$ indexes individual horizons. 
    
\begin{figure}[htbp]
        \centering
        \begin{tikzpicture}[scale=1]
            
            \node[inner sep=0pt] (img) at (0,0)
            {\includegraphics[width=0.50\textwidth, trim=0cm 6cm 0cm 3cm, clip]{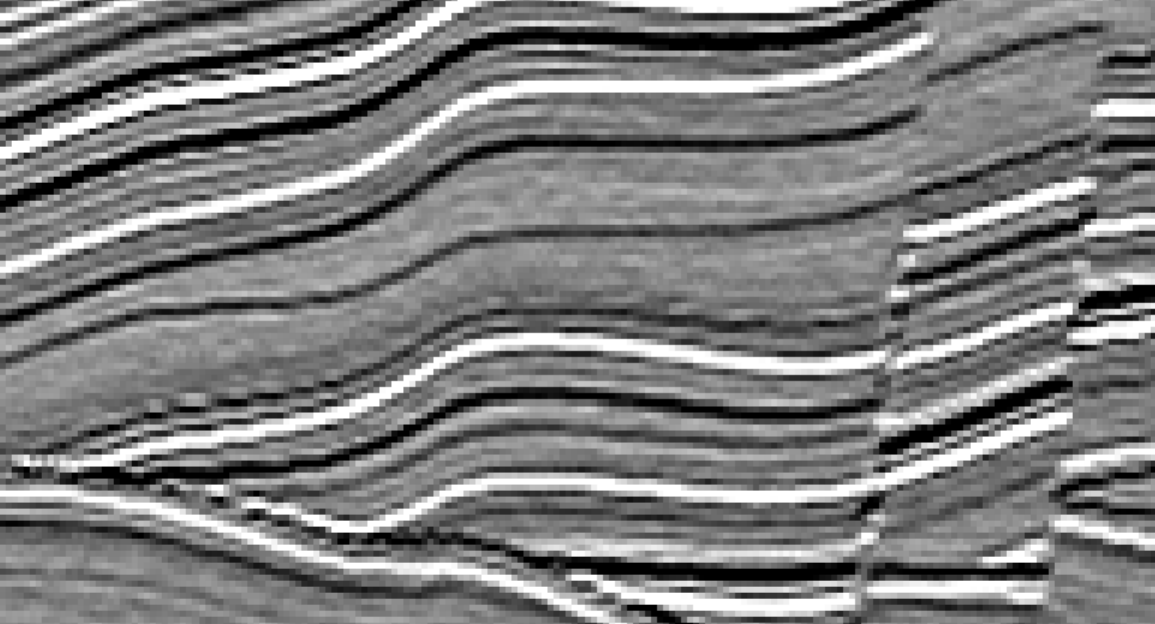}};	
            
            \path (img.south west) -- (img.north west) coordinate[pos=0.41] (hs1);
            
            \draw [dotted, line width=0.7mm, green, opacity=0.8] plot [smooth] coordinates {(hs1) (-4, -0.2) (-3, 0.) (-2, 0.05) (-1, 0.45) (0, 0.6) (1, 0.6) (2, 0.7) (2.55, 0.9) (2.75, 1.75) (3, 1.9) (4.15, 2.2)};

            \path (img.south west) -- (img.north west) coordinate[pos=0.25] (hs2);
            \path (img.south east) -- (img.north east) coordinate[pos=0.825] (he2);
            
            \draw [dotted, line width=0.7mm, orange, opacity=0.8] plot [smooth] coordinates {(hs2) (-4, -1.2) (-3, -1.05) (-2, -0.85) (-1, -0.4) (0, -0.25) (1, -0.4) (2, -0.45) (2.4, -0.3) (2.6, 0.5) (3.9, 1) (4.15, 1.55) (he2)};

            \draw (-4, 0.1) node[opacity=0.8] {$\textcolor{green}{k}$};
            \draw (-4, -0.9) node[opacity=0.8] {$\textcolor{orange}{k'}$};

            \draw[thick, red] (0, -2.44) -- (0, 2.43);
            \draw (-0.3, 0.9) node[red] {$\mathbf{p}_1$};

            \draw[thick, blue] (1, -2.44) -- (1, 2.43);
            \draw (0.7, 0.9) node[blue] {$\mathbf{p}_2$};
            
            \node[fill=white, draw=black, rounded corners, opacity=0.9] at (-1,1.75) 
            {\large $g_{\theta}(\textcolor{red}{\mathbf{p}_1}, \nu_{\textcolor{ForestGreen}{k}}(\textcolor{red}{\mathbf{p}_1})) = g_{\theta}(\textcolor{blue}{\mathbf{p}_2}, \nu_{\textcolor{ForestGreen}{k}}(\textcolor{blue}{\mathbf{p}_2}))$};
            
            \node[fill=white, draw=black, rounded corners, opacity=0.9] at (-1,-1.75) 
            {\large $g_{\theta}(\textcolor{red}{\mathbf{p}_1}, \nu_{\textcolor{ForestGreen}{k}}(\textcolor{red}{\mathbf{p}_1})) \neq g_{\theta}(\textcolor{red}{\mathbf{p}_1}, \nu_{\textcolor{orange}{k'}}(\textcolor{red}{\mathbf{p}_1}))$};

        \end{tikzpicture}
        \caption{A 2D slice of a synthetic dataset with two horizons (green and orange dotted lines). Our method promotes similar representations for voxels belonging to the same horizon, while encouraging separation for representations across different depths along the same trace (Eq.~\ref{eq:nu_k_def}).}
        \label{fig:method_illustration}
    \end{figure}
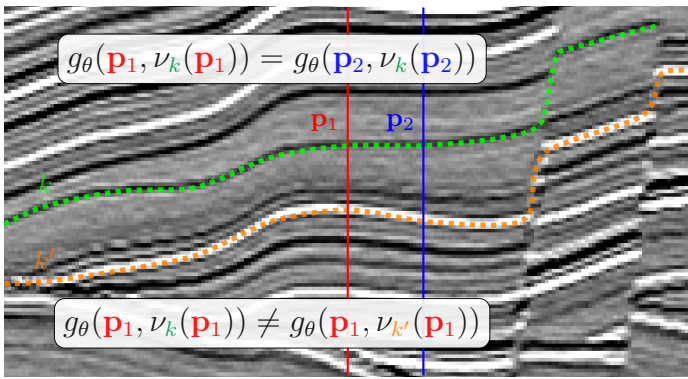

    We aim to learn a mapping $g_\theta : S \times T \rightarrow \mathbb{R}^N$ such that voxels lying on the same horizon share similar embeddings (see Fig.~\ref{fig:method_illustration}):
    \begin{equation}
         \exists k, t_1 = \nu_k(\mathbf{p}_1) \wedge t_2 = \nu_k(\mathbf{p}_2) \Rightarrow g_{\theta}(\mathbf{p}_1, t_1) \approx g_{\theta}(\mathbf{p}_2 , t_2)
        \label{eq:nu_k_def}
    \end{equation}

    Since $\nu_k$ is unknown, we use signal-based local correspondences as priors to learn voxel-wise representations aligned with local approximations of $\nu_k$. We implement $g_\theta$ with a 1D encoder $f_\theta : \mathbb{R}^T \rightarrow \mathbb{R}^{T \times N}$, so that $g_\theta(\mathbf{p},t) = f_\theta(\mathbf{x}_{\mathbf{p}})_t = \mathbf{Y}_{\mathbf{p},t}$.
		
    \subsection{Contrastive Learning}

    We use a trace-to-trace flow $\phi_{\mathbf{p}_i, \mathbf{p}_j}: \{1, \dots, T\} \rightarrow [1,T]$ to map a time index from trace $\mathbf{p}_i$ to its corresponding index on trace $\mathbf{p}_j$. This local correspondence can be expressed as follows:
    \begin{equation}
        \mathbf{x}_{\mathbf{p}_i, t} \approx \mathbf{x}_{\mathbf{p}_j, \phi_{\mathbf{p}_i, \mathbf{p}_j}(t)} \Rightarrow \mathbf{Y}_{\mathbf{p}_i, t} \approx \mathbf{Y}_{\mathbf{p}_j, \phi_{\mathbf{p}_i, \mathbf{p}_j}(t)}
    \end{equation}
    For an anchor voxel $(\mathbf{p},t)$, a positive sample is defined by the mapped voxel $(\mathbf{p}',\phi_{\mathbf{p},\mathbf{p}'}(t))$ for $\mathbf{p}'$ in a local neighborhood $\mathcal{B}(\mathbf{p}, t)$. Negatives are taken as other depth samples along the anchor trace $\mathbf{p}$, encouraging separation between nearby horizons. Following InfoNCE~\cite{oord2019representationlearningcontrastivepredictive}, the loss at depth $t$ is:
    \begin{equation}
        L(\mathbf{p}, t) = \sum_{\mathbf{p}' \in \mathcal{B}(\mathbf{p}, t)} -\ln \left[ \frac{\exp(s_{\mathbf{p}, \mathbf{p}', t}^+)}{\exp(s_{\mathbf{p}, \mathbf{p}', t}^+) + \sum_{t'\neq t}\exp(s_{\mathbf{p}, t, t'}^-)} \right]
        \label{eq:infonce}
    \end{equation}
    where $s_{\mathbf{p}, \mathbf{p}', t}^+ = \cosim(\mathbf{Y}_{\mathbf{p}, t}, \mathbf{Y}_{\mathbf{p}', \phi_{\mathbf{p}, \mathbf{p}'}(t)}) / \tau$ denotes the positive pair similarity, $s_{\mathbf{p}, t, t'}^- = \cosim(\mathbf{Y}_{\mathbf{p}, t}, \mathbf{Y}_{\mathbf{p}, t'})/ \tau$ denotes the negative pair similarity, $\cosim$ is the cosine similarity, and $\tau > 0$ is a temperature hyperparameter. The loss is averaged over sampled anchors at different depths and positions.

    We handle non-integer $\phi(t)$ by linear interpolation. The main challenge is to restrict training to neighborhoods $\mathcal{B}$ where flow-based correspondences are reliable. Otherwise, false positives may align different horizons. Next, we describe how $\phi$ and $\mathcal{B}$ are computed.
    
    \subsection{Flow $\phi$ Estimation}
    
    \subsubsection{Seismic Reflector Slope}
	
    Numerous methods have been proposed to estimate seismic reflector slopes~\cite{li2021flowNet, hale2006dipCorr, gijs2003structurTensor, lou2022dipDeep}. We use the structure tensor-based method~\cite{gijs2003structurTensor}, from which we derive $\phi_{\mathbf{p}, \mathbf{p} + \mathbf{v}}$ and an associated confidence measure $c_{\mathbf{p}, \mathbf{p} + \mathbf{v}}: \{1, \dots, T\} \rightarrow [0, 1]$. Here, $\mathbf{v}\in\{(\pm 1, 0), (0, \pm 1)\}$ connects 4-neighbor traces on the 2D grid $S$. For each neighboring direction $\mathbf{v}$, the estimated reflector slope $u_{\mathbf{v}}(\mathbf{p}, t)$ from $\mathbf{p}$ to $\mathbf{p} + \mathbf{v}$ at depth $t$ is converted into a local trace-to-trace flow:
    \begin{equation}
        \phi_{\mathbf{p}, \mathbf{p} + \mathbf{v}}(t) = t + u_{\mathbf{v}}(\mathbf{p},t)
    \end{equation} 
    A deformation field between two given points $\mathbf{p}_1$ and $\mathbf{p}_n$ is obtained by composing local deformation fields:
    \begin{equation}
        \phi_{\mathbf{p}_1, \mathbf{p}_n}(t) = (\phi_{\mathbf{p}_{n-1}, \mathbf{p}_n} \circ \cdots \circ \phi_{\mathbf{p}_2, \mathbf{p}_3} \circ  \phi_{\mathbf{p}_1, \mathbf{p}_2})(t)
    \end{equation}
    where $\mathbf{p}_i - \mathbf{p}_{i-1} \in \{(\pm 1, 0),(0, \pm 1)\}$.
    
    The path $(\mathbf{p}_1,\ldots,\mathbf{p}_n)$ is not unique and we favor paths that avoid noisy, faulted, or chaotic areas. Path quality is measured using the confidence index $c$, which is low where no dominant orientation exists (typically near faults or in chaotic regions). However, fault delineation using this method may be incomplete, potentially resulting in inaccurate neighborhood estimates near faults. We optionally combine $c$ with a deep-learning fault mask (e.g.,~\cite{yu2023faultSurvey}) from a pretrained model. Masked voxels are assigned zero confidence. The mask is not intended to improve horizon tracking near faults, but to prevent ambiguous fault-crossing correspondences from being selected as contrastive positives (results are also reported without it).

    \subsubsection{Neighborhood Construction}

    We model the confidence $\mu$ along a path $P = (\mathbf{p}_0, \mathbf{p}_1, \cdots, \mathbf{p}_{n-1}, \mathbf{p}_n)$ at depth $t$ using a multiplicative formulation:
    \begin{equation}
        \mu_t(P) = \prod_{i=0}^{n-1} c_{\mathbf{p}_i, \mathbf{p}_{i+1}}(\phi_{\mathbf{p}_0, \mathbf{p}_{i}}(t))
    \end{equation}
    By applying a logarithmic transform, each transition $(\mathbf{p}_i, \phi_{\mathbf{p}_0, \mathbf{p}_{i}}(t)) \rightarrow (\mathbf{p}_{i+1}, \phi_{\mathbf{p}_0, \mathbf{p}_{i+1}}(t))$ is associated with the local cost:
    \begin{equation}
        w_{\mathbf{p}_i, \mathbf{p}_{i+1}}(\phi_{\mathbf{p}_0, \mathbf{p}_{i}}(t)) = -\ln(\max(\epsilon, c_{\mathbf{p}_i, \mathbf{p}_{i+1}}(\phi_{\mathbf{p}_0, \mathbf{p}_{i}}(t))))
    \end{equation}
    The total path cost is then given by:
    \begin{equation}
        d_{t}(P) = \sum_{i=0}^{n-1} w_{\mathbf{p}_i, \mathbf{p}_{i+1}}(\phi_{\mathbf{p}_0, \mathbf{p}_{i}}(t))
    \end{equation}
    We define the neighborhood $\mathcal{B}$ at depth $t$ and threshold $\delta$ as the set of traces that can be reached from $\mathbf{p}_0$ through at least one path whose cumulative cost does not exceed $\delta$:
    \begin{equation}
        \mathcal{B}(\mathbf{p}_0,t) = \{\mathbf{p} \in S | \exists P\in \mathcal{P} (\mathbf{p}_0, \mathbf{p}), d_t(P) \leq \delta \}
    \end{equation}
    where $\mathcal{P}(\mathbf{p}_0,\mathbf{p})$ denotes the set of spatial paths from $\mathbf{p}_0$ to $\mathbf{p}$.

    In practice, this set is approximated by a front-propagation procedure to limit computational cost. This approximation may underestimate $\mathcal{B}(\mathbf{p}_0,t)$. The neighborhood is constructed incrementally from the anchor $(\mathbf{p}_0,t)$, initialized as $\mathcal{B}(\mathbf{p}_0,t) = \{\mathbf{p}_0\}$. Let $P_{\mathbf{p}}$ denote the path assigned to $\mathbf{p}$ when $\mathbf{p}$ is added to $\mathcal{B}(\mathbf{p}_0,t)$. A neighboring trace $\mathbf{p}+\mathbf{v}$, with $\mathbf{v}\in\{(\pm1,0), (0,\pm1)\}$ and $\mathbf{p} + \mathbf{v} \notin \mathcal{B}(\mathbf{p}_0, t)$, defines a candidate extension through the path $P' = (P_{\mathbf{p}},\mathbf{p}+\mathbf{v})$. Among all such candidate extensions, we select the one with minimum cumulative cost and add its endpoint $\mathbf{p}+\mathbf{v}$ to $\mathcal{B}(\mathbf{p}_0,t)$ if $d_t(P') \leq \delta$. The propagation proceeds by expanding candidate extensions in increasing order of cumulative cost. In the contrastive loss, the anchor trace itself is excluded from $\mathcal{B}(\mathbf{p}_0,t)$ when forming positive pairs.
    
    \section{Results}

    \subsection{Evaluation Protocol}

    After training on a given 3D seismic volume, each voxel is associated with an $N$-dimensional embedding. To reconstruct a horizon, we start from a seed point $(\mathbf{p}, t)$ and use its embedding $\mathbf{Y}_{\mathbf{p}, t}$ as reference. The predicted surface is obtained by selecting, for each $\mathbf{p}'\in S$, the depth that maximizes cosine similarity:
    \begin{equation}
        \tilde{\nu}_{\mathbf{p}, t}(\mathbf{p}') = \argmax_{t' \in \{1, \dots, T\}} \cosim\left(\mathbf{Y}_{\mathbf{p}, t}, \mathbf{Y}_{\mathbf{p}', t'} \right)
    \end{equation}
    
    For evaluation, we choose one seed point in the middle of the ground-truth horizon $\nu_k$, where the seed is assumed to lie on the sought horizon, and report the mean absolute error (MAE):
    \begin{align}
        \text{MAE}(\nu_k, \tilde{\nu}_{\mathbf{p}, t}) = \frac{1}{|S|} \sum_{\mathbf{p}' \in S} \left|\nu_k(\mathbf{p}') - \tilde{\nu}_{\mathbf{p}, t}(\mathbf{p}') \right|
    \end{align}
    For all methods, MAE is computed on the same voxels outside the fault mask. This protocol evaluates horizon propagation across faults. The objective is to recover the same seeded horizon in interpretable regions on both sides of discontinuities, not to predict horizon positions inside fault zones where reflector correspondences may be ambiguous or ill-defined. 
    
    \begin{figure}[!htbp]
        \centering
        \begin{tikzpicture}
        \begin{axis}[
        xmin=0,xmax=4,
        ymin=-1,ymax=1,
        zmin=-1,zmax=1,
        xlabel=$x$,ylabel=$y$,zlabel=$z$,
        view={35}{30},
        hide axis
        ]
        \path (axis cs:0,0,0) coordinate (O) (axis cs:1,0,0) coordinate (X)  
        (axis cs:0,1,0) coordinate (Y) (axis cs:0,0,1) coordinate (Z)
        (axis cs:2,1,0) coordinate (Px)  (axis cs:0,0,0) coordinate (Py) (axis cs:2,0,-1) coordinate (Pz);
        \end{axis}
         \begin{scope}[x={($(X)-(O)$)},y={($(Y)-(O)$)},z={($(Z)-(O)$)},
            canvas is xz plane at y=0,transform shape]
          \path (Px) node{\includegraphics[width=4cm,height=2cm]{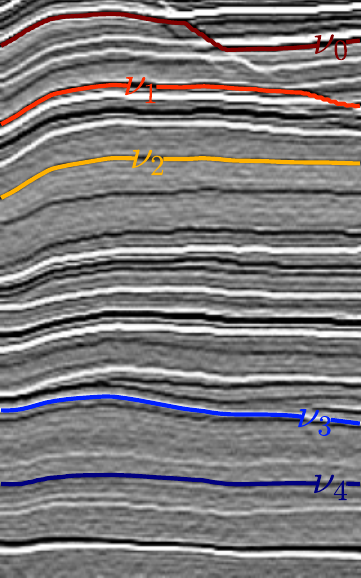}};
         \end{scope}
         \begin{scope}[x={($(X)-(O)$)},y={($(Y)-(O)$)},z={($(Z)-(O)$)},
            canvas is yz plane at x=0,transform shape]
          \path (Py) node{\includegraphics[width=2cm,height=2cm]{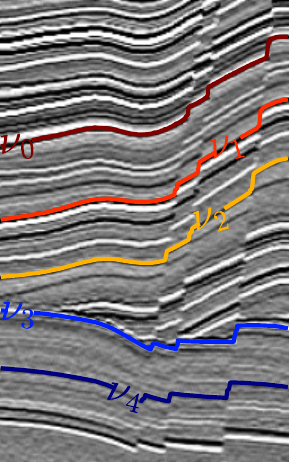}};
         \end{scope}
        \end{tikzpicture}
        
        \caption{Representative inline and crossline sections of the synthetic 3D seismic cube with the five ground-truth horizons.}
        \label{fig:syn_dataset_3D}
    \end{figure}

    \subsection{Implementation Details}

    The 3D seismic cubes are normalized using a z-score. We train a 1D U-Net~\cite{ronneberger2015unet} (output dimension $N = 32$ for each voxel) using the Adam optimizer (learning rate $10^{-3}$) for 20 epochs. The InfoNCE temperature is set to $\tau = 0.7$. For within-trace negatives, we use a thresholded negative similarity:
    \begin{equation}
        \tilde{s}_{\mathbf{p}, t, t'}^{-} =
    \begin{cases}
    \cosim(\mathbf{Y}_{\mathbf{p}, t}, \mathbf{Y}_{\mathbf{p}, t'}) /\tau,
    & \cosim(\mathbf{Y}_{\mathbf{p}, t},\mathbf{Y}_{\mathbf{p}, t'}) \geq \alpha,\\
    0 & \text{otherwise}
    \end{cases}
    \end{equation}
    with $\alpha=0.5$ unless stated otherwise. Negative pairs with cosine similarity below $\alpha$ are assigned a constant logit equal to zero, so that only hard negatives above the threshold are emphasized. Neighborhoods $\mathcal{B}(\mathbf{p}, t)$ are computed on the fly within a $20 \times 20$ spatial window using a graph-distance threshold $\delta = -\log(0.7)$ and $\epsilon = 10^{-5}$. Anchor voxels $(\mathbf{p}, t)$ are sampled uniformly along the depth dimension. All experiments were conducted on an  80 GB NVIDIA A100 GPU.

    \subsection{Datasets}

    We evaluate the proposed method on the F3 dataset~\cite{silva2019netherlands} using seven interpreted ground-truth horizons. In addition, we introduce a synthetic dataset generated with \textit{Synthoseis}~\cite{merrifield2022synGen}. This synthetic volume of dimensions $L = W = 160$ and $T = 256$ contains five major horizons disrupted by three significant faults. This synthetic dataset is used for computationally intensive experiments, as its smaller size makes such evaluations more tractable. A depiction of the 3D volume and its five ground-truth horizons is shown in Fig.~\ref{fig:syn_dataset_3D}. 
    
    \subsection{Benchmark}

    We compare our method to three representative baselines: \begin{enumerate*}[label=(\roman*)]
        \item a correlation-based propagator with a single seed (Corr), using a window size of $10$ for the synthetic dataset and $12$ for F3, selected to provide sufficient temporal context for the expected vertical displacements~\cite{guillon2010Corr},
        \item an RGT-based approach~\cite{guillon2013geotime}, and 
        \item Geo2.5D~\cite{thouvenot2025geo25d}, which combines texture and signal in a semi-supervised setting.
    \end{enumerate*}
    The first two baselines are fully unsupervised, while Geo2.5D uses a single labeled 2D slice per dataset. Results are reported in Table~\ref{tab:benchmark_method}. Reported MAE values are averaged over the horizons.
    \begin{table}[htbp]
        \centering
        \caption{MAE comparison. Ours$^{-}$ excludes the fault mask prior; Ours$^{+}$ includes it.}
        \label{tab:benchmark_method}
        \setlength{\tabcolsep}{4pt}
        \begin{tabular}{lcc|cc|c}
            \toprule
            & \multicolumn{2}{c|}{\textbf{Unsupervised}} & \multicolumn{2}{c|}{\textbf{Self-Supervised}} & \textbf{Semi-Supervised} \\
            & Corr~\cite{guillon2010Corr} & RGT~\cite{guillon2013geotime} & Ours$^-$ & Ours$^+$ & Geo2.5D~\cite{thouvenot2025geo25d} \\
            \midrule
            Synthetic & $5.83$ & $5.72$ & $0.43$ & $0.40$ & $\textbf{0.35}$ \\
            F3        & $7.78$ & $4.11$ & $2.24$ & $\textbf{1.99}$ & $2.68$ \\
            \bottomrule
        \end{tabular}
    \end{table}

\begin{figure}[!htbp]
        \centering
        \subfloat[]{\includegraphics[width=0.98\linewidth]{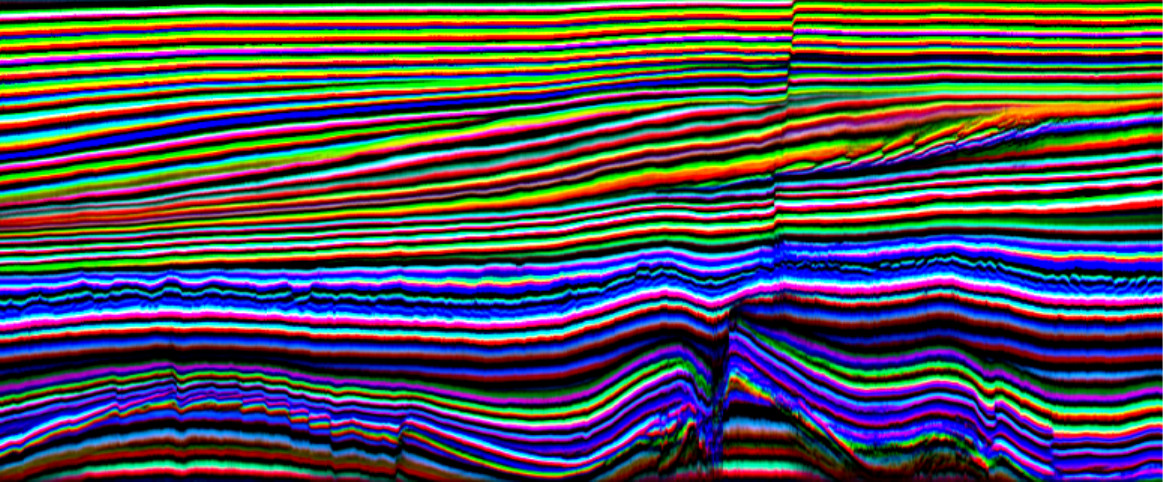}
        \label{fig:pca_f3}}
        \vspace{0.5em}
        \subfloat[]{\includegraphics[width=0.98\linewidth]{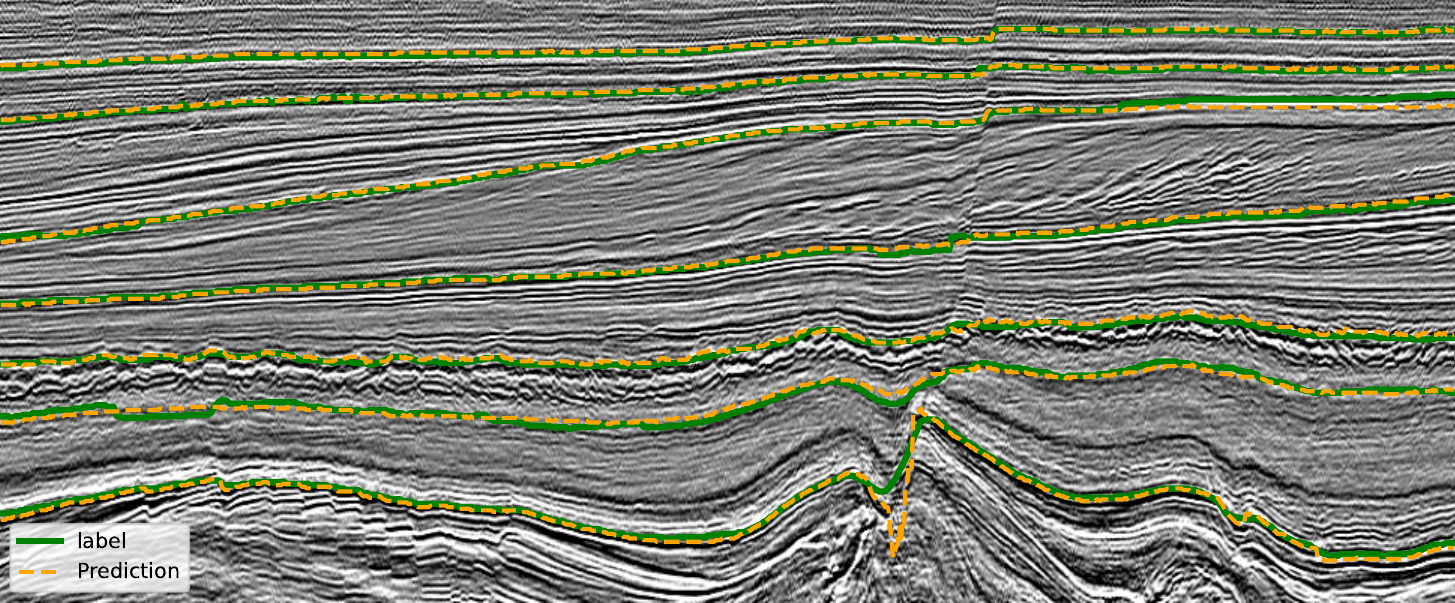}
        \label{fig:h_f3}}
        
        \caption{Predictions on the F3 dataset obtained with a model trained using a fault mask prior, in an area affected by faults. 
        (a) PCA projection of the $N$-dimensional representation to RGB space. 
        (b) Predicted horizons versus ground truth.}
        \label{fig:f3_benchmark}
    \end{figure}

\begin{figure}[htbp]
        \centering
        \subfloat[]{\includegraphics[width=0.23\textwidth]{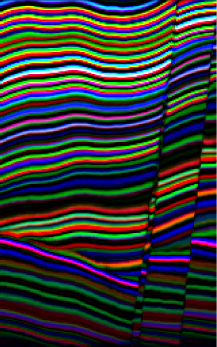}%
        \label{fig:pca_syn}}
        \hfill
        \subfloat[]{\includegraphics[width=0.23\textwidth]{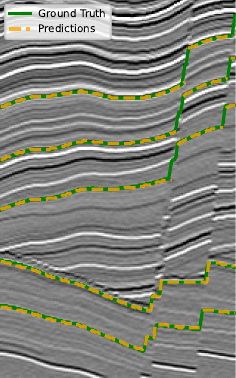}%
        \label{fig:h_syn}}
        \caption{Predictions on the synthetic dataset illustrate horizon propagation in the presence of three major faults.
        (a) PCA projection of the $N$-dimensional representation to RGB space.
        (b) Predicted horizons versus ground truth.}
        \label{fig:syn_benchmark}
    \end{figure}

    Geo2.5D achieves strong accuracy with a single labeled slice, illustrating the benefit of semi-supervision. Without horizon labels, our method substantially improves over correlation and RGT baselines and remains competitive with Geo2.5D. The optional fault mask further improves performance by suppressing unreliable positive pairs, while Ours$^-$ shows that the method remains effective without this prior. Fig.~\ref{fig:f3_benchmark} and Fig.~\ref{fig:syn_benchmark} visualize the learned embeddings with principal component analysis (PCA) and the corresponding horizon reconstructions in areas affected by faults.

    \subsection{Ablation and Sensitivity Analysis}
    
     Ablations are performed on the synthetic dataset using the default configuration: 1D U-Net, $N=32$, and $\delta = -\log(0.7)$, unless stated otherwise.

    \subsubsection{Network Architecture}

    In this section, we evaluate the influence of the network architecture and compare a 1D UNET~\cite{ronneberger2015unet}, a 1D UNET with a ResNet backbone~\cite{diakogiannis2020uRes}, and a 1D version of the UNETR hybrid architecture~\cite{hatamizadeh2021unetrtransformers3dmedical}. As shown in Table~\ref{tab:benchmark_models_full}, the 1D U-Net matches the best MAE while using fewer parameters, and is therefore used in all subsequent experiments.

    \begin{table}[htbp]
        \centering
        \caption{MAE comparison on three architectures.}
        \label{tab:benchmark_models_full}
        \begin{tabular}{lccc}
            \toprule
            Model & 1D U-Net & 1D UNETR & 1D U-Net ResNet \\
            \midrule
            MAE $\downarrow$ & $\textbf{0.40}$ & $\textbf{0.40}$ & 0.63 \\
            Parameters (M) & 17.1 & 19.7 & 21.6 \\
            \bottomrule
        \end{tabular}
    \end{table}

    \subsubsection{Seed Sensitivity}

    To assess the influence of the seed location, we repeat horizon tracking from 225 different seeds sampled on a regular $15 \times 15$ spatial grid (outside the fault mask). For each horizon $\nu_k$, each seed $(\mathbf{p}_i, \nu_k(\mathbf{p}_i))$ produces one propagated surface $\tilde{\nu}_{\mathbf{p}_i, \nu_k(\mathbf{p}_i)}$. We summarize performance in Fig.~\ref{fig:benchmark_seed_sensitivity} by reporting the mean and standard deviation of the MAE. The relatively large standard deviation reflects genuine spatial heterogeneity. Seeds near faults or low-confidence regions are more likely to produce partial tracking failures.
    
    \subsubsection{Embedding Dimension}

    The dimensionality of the representation space $N$ may need to increase with the complexity or the number of horizons. To investigate this relationship, we evaluate embedding dimension $N \in \{2, 4, 8, \dots, 1024\}$. We report the results in Fig.~\ref{fig:depth_embedding}. Low-dimensional representations ($N = 2$ and $N = 4$) yield poor performance, whereas higher dimensions result in more stable outcomes. Performance stabilizes once sufficient embedding dimension is selected.

    \subsubsection{Angular Relaxation}
    
    We evaluate the cosine threshold $\alpha \in \{0.1, \dots, 0.9\}$ used in the angular-relaxed negative similarity. Fig.~\ref{fig:benchmark_alpha} shows a U-shaped MAE trend, with best performance for moderate thresholds $\alpha \in (0.4, 0.6)$. Standard InfoNCE without angular relaxation yields a higher MAE of $0.58$.
      
    \subsubsection{Discontinuity Handling}

\begin{figure*}[htbp]
        \subfloat[]{\includegraphics[width=0.24\textwidth]{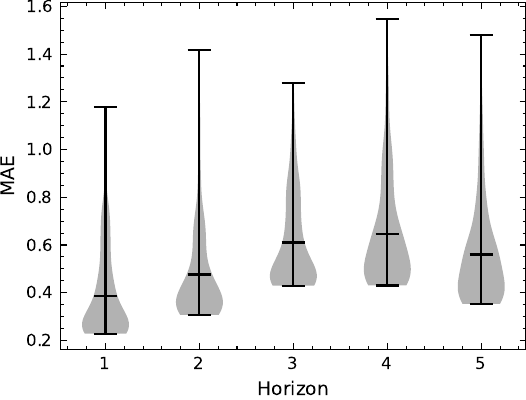}%
        \label{fig:benchmark_seed_sensitivity}}
        \hfill
        \subfloat[]{\includegraphics[width=0.24\textwidth]{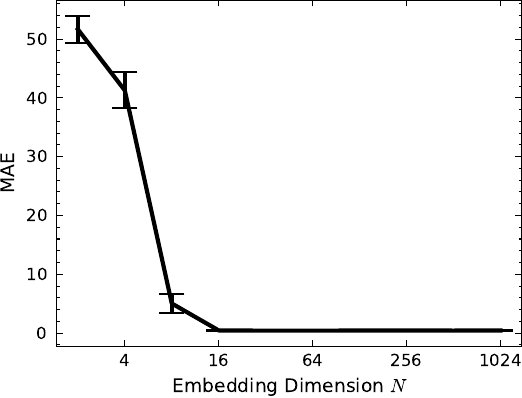}%
        \label{fig:depth_embedding}}
        \hfill
        \subfloat[]{\includegraphics[width=0.24\textwidth]{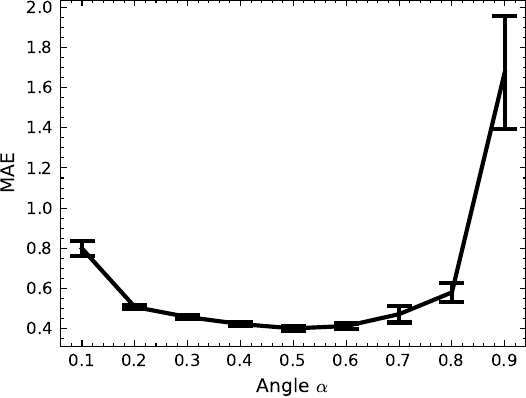}%
        \label{fig:benchmark_alpha}}
        \hfill
        \subfloat[]{\includegraphics[width=0.24\textwidth]{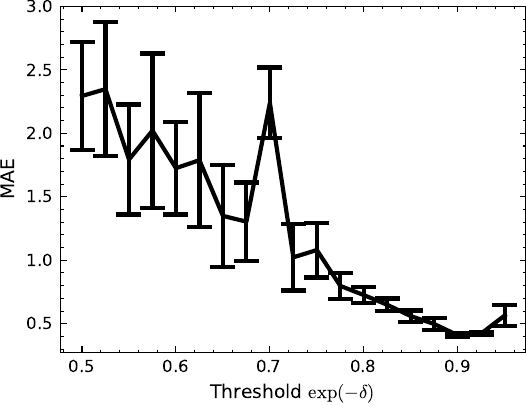}%
        \label{fig:benchmark_coherence}}        

        \caption{Ablation studies on the synthetic dataset (results averaged over five runs). 
        (a) Seed sensitivity: distribution of MAE over $225$ seed locations sampled on a regular spatial grid for each horizon (mean indicated).
        (b) Impact of embedding dimension $N$: values greater than or equal to 16 show improved performance.
        (c) Impact of angular relaxation: values $(0.3, 0.7)$ show improved stability and horizon propagation.
        (d) Impact of the path confidence threshold $\exp(-\delta)$ without a fault mask: higher values improve horizon propagation.
        Error bars in (b)–(d) denote mean $\pm$ standard error (SE).}
        
    \end{figure*}

    We first study $\delta$ without a fault mask, using only the confidence to prevent unreliable local correspondences. As shown in Fig.~\ref{fig:benchmark_coherence}, we evaluate $\exp(-\delta) \in [0.5, 0.95]$. Values in $[0.85, 0.95]$ yield the best performance, while lower values degrade results, highlighting the role of $\delta$ in selecting reliable positive pairs across structurally complex areas.

    \section{Conclusion} 
    
    We proposed a self-supervised contrastive framework for horizon propagation in 3D seismic volumes, using signal-based local correspondences to learn voxel-wise embeddings. Experiments on a synthetic faulted benchmark and the public F3 dataset show that the proposed method improves over unsupervised baselines for horizon propagation across faulted regions and remains competitive with a semi-supervised approach using a single labeled slice. Future work will focus on relaxing the locality assumption used in neighborhood construction. In particular, extending the graph-based search beyond local windows, while controlling the reliability of long-range correspondences, could improve global consistency in complex areas.

    \section*{Acknowledgment}

    This work was supported by TotalEnergies and the French National Association for Research and Technology (ANRT) through a CIFRE Ph.D. fellowship. 
    
    The authors declare that this work is related to a patent application filed by their institution.

    \bibliographystyle{IEEEtran}
    \bibliography{main}

@ARTICLE{Wang2023segFormer,
	author={Wang, Zhiguo and Wang, Qiannan and Yang, Yang and Liu, Naihao and Chen, Yumin and Gao, Jinghuai},
	journal={IEEE Transactions on Geoscience and Remote Sensing}, 
	title={Seismic Facies Segmentation via a {SegFormer}-Based Specific Encoder–Decoder–Hypercolumns Scheme}, 
	year={2023},
	volume={61},
	number={},
	pages={1-11},
	doi={10.1109/TGRS.2023.3244037}}

@ARTICLE{Li2023conss,
	author={Li, Kewen and Liu, Wenlong and Dou, Yimin and Xu, Zhifeng and Duan, Hongjie and Jing, Ruilin},
	journal={IEEE Journal of Selected Topics in Applied Earth Observations and Remote Sensing}, 
	title={{CONSS}: Contrastive Learning Method for Semisupervised Seismic Facies Classification}, 
	year={2023},
	volume={16},
	number={},
	pages={7838-7849},
	doi={10.1109/JSTARS.2023.3308754}}

@article{abubakar2022rgtFlow,
	author = { Aria Abubakar  and  Haibin Di  and  Zhun Li  and  Hiren Maniar  and  Tao Zhao },
	title = {An artificial intelligence workflow for horizon volume generation from 3-D seismic data},
	journal = {The Leading Edge},
	volume = {43},
	number = {4},
	pages = {235--243},
	year = {2024},
	doi = {10.1190/tle43040235.1},
	eprint = {https://doi.org/10.1190/tle43040235.1}
}

@inproceedings{li2021flowNet,
  title={Seismic flownet: Using optical flow field for dense horizon interpretation},
  author={Li, Z and Abubakar, A},
  booktitle={82nd EAGE annual conference \& exhibition},
  volume={2021},
  number={1},
  pages={1--5},
  year={2021},
  organization={European Association of Geoscientists \& Engineers}
}

@article{guillon2010Corr,
    author = {Gallon, Jonathan and Guillon, Sebastien and Jobard, Bruno and Barucq, Helene and Keskes, Noomane},
    year = {2010},
    month = {05},
    pages = {37--44},
    title = {Slimming Brick Cache Strategies for Seismic Horizon Propagation Algorithms},
    journal = {Eurographics/IEEE VGTC on Volume Graphics},
    doi = {10.2312/VG/VG10/037-044}
}

@article{Lomask2006rgt,
	author = {Jesse Lomask and Antoine Guitton and Sergey Fomel and Jon Claerbout and Alejandro A. Valenciano},
	title = {Flattening without picking},
	journal = {GEOPHYSICS},
	volume = {71},
	number = {4},
	pages = {P13--P20},
	year = {2006},
	doi = {10.1190/1.2210848},
	eprint = {https://doi.org/10.1190/1.2210848}
}

@Article{su2023dtw,
	AUTHOR = {Su, Mingjun and Qian, Feng and Cui, Shengkai and Yuan, Cheng and Cui, Xiangli},
	TITLE = {Research on a 3-D Seismic Horizon Automatic-Tracking Method Based on Corrugated Global Diffusion},
	JOURNAL = {Applied Sciences},
	VOLUME = {13},
	YEAR = {2023},
	NUMBER = {10},
	ARTICLE-NUMBER = {6155},
	ISSN = {2076-3417},
	DOI = {10.3390/app13106155}
}

@ARTICLE{dramsch2022voxelmorph4D,
	author={Dramsch, Jesper Sören and Christensen, Anders Nymark and MacBeth, Colin and Lüthje, Mikael},
	journal={IEEE Transactions on Geoscience and Remote Sensing}, 
	title={Deep Unsupervised 4-D Seismic 3-D Time-Shift Estimation With Convolutional Neural Networks}, 
	year={2022},
	volume={60},
	number={},
	pages={1--16},
	doi={10.1109/TGRS.2021.3081516}}

@article{merrifield2022synGen,
	author = {Tom P. Merrifield and Donald P. Griffith and S. Ahmad Zamanian and Stephane Gesbert and Satyakee Sen and Jorge De La Torre Guzman and R. David Potter and Henning Kuehl},
	title = {Synthetic seismic data for training deep learning networks},
	journal = {Interpretation},
	volume = {10},
	number = {3},
	pages = {SE31--SE39},
	year = {2022},
	doi = {10.1190/INT-2021-0193.1}}

@article{oord2019representationlearningcontrastivepredictive,
  title={Representation learning with contrastive predictive coding},
  author={Oord, Aaron van den and Li, Yazhe and Vinyals, Oriol},
  journal={arXiv preprint arXiv:1807.03748},
  year={2018}
}

@article{silva2019netherlands,
  title={Netherlands dataset: A new public dataset for machine learning in seismic interpretation},
  author={Silva, Reinaldo Mozart and Baroni, Lais and Ferreira, Rodrigo S and Civitarese, Daniel and Szwarcman, Daniela and Brazil, Emilio Vital},
  journal={arXiv preprint arXiv:1904.00770},
  year={2019}
}

@proceedings{hale2006dipCorr,
    author = {Hale, Dave},
    title = {Fast Local Cross-correlations of Images},
    volume = {2006 SEG Annual Meeting},
    series = {SEG International Exposition and Annual Meeting},
    pages = {SEG-2006-3160},
    year = {2006},
    month = {10},
    eprint = {https://onepetro.org/SEGAM/proceedings-pdf/SEG06/SEG06/SEG-2006-3160/1845228/seg-2006-3160.pdf},
}

@article{gijs2003structurTensor,
	author = {Gijs C. Fehmers and Christian F. W. Höcker},
	title = {Fast structural interpretation with structure‐oriented filtering},
	journal = {GEOPHYSICS},
	volume = {68},
	number = {4},
	pages = {1286--1293},
	year = {2003},
	doi = {10.1190/1.1598121},
	eprint = {https://doi.org/10.1190/1.1598121}
}

@article{lou2022dipDeep,
	title = {Seismic volumetric dip estimation via a supervised deep learning model by integrating realistic synthetic data sets},
	journal = {Journal of Petroleum Science and Engineering},
	volume = {218},
	pages = {111021},
	year = {2022},
	issn = {0920-4105},
	doi = {https://doi.org/10.1016/j.petrol.2022.111021},
	author = {Yihuai Lou and Shizhen Li and Naihao Liu and Rongchang Liu}
}

@article{pham2025seisBERT,
	author = {Nam Pham  and  Haibin Di  and  Tao Zhao  and  Aria Abubakar },
	title = {SeisBERT: A pretrained seismic image representation model for seismic data interpretation},
	journal = {The Leading Edge},
	volume = {44},
	number = {2},
	pages = {96--106},
	year = {2025},
	doi = {10.1190/tle44020096.1}
}

@InProceedings{goncharov2023voxel2vex,
  title={vox2vec: A framework for self-supervised contrastive learning of voxel-level representations in medical images},
  author={Goncharov, Mikhail and Soboleva, Vera and Kurmukov, Anvar and Pisov, Maxim and Belyaev, Mikhail},
  booktitle={International Conference on Medical Image Computing and Computer-Assisted Intervention},
  pages={605--614},
  organization={Springer},
  year={2023}
}

@InProceedings{xie2021pixelCl,
	author    = {Xie, Zhenda and Lin, Yutong and Zhang, Zheng and Cao, Yue and Lin, Stephen and Hu, Han},
	title     = {Propagate Yourself: Exploring Pixel-Level Consistency for Unsupervised Visual Representation Learning},
	booktitle = {Proceedings of the IEEE/CVF Conference on Computer Vision and Pattern Recognition (CVPR)},
	month     = {June},
	year      = {2021},
	pages     = {16684--16693}
}

@inproceedings{ronneberger2015unet,
  title={{U-Net}: Convolutional networks for biomedical image segmentation},
  author={Ronneberger, Olaf and Fischer, Philipp and Brox, Thomas},
  booktitle={International Conference on Medical image computing and computer-assisted intervention},
  pages={234--241},
  year={2015},
  organization={Springer}
}

@article{diakogiannis2020uRes,
   title={{ResUNet}-a: A deep learning framework for semantic segmentation of remotely sensed data},
   volume={162},
   ISSN={0924-2716},
   DOI={10.1016/j.isprsjprs.2020.01.013},
   journal={ISPRS Journal of Photogrammetry and Remote Sensing},
   publisher={Elsevier BV},
   author={Diakogiannis, Foivos I. and Waldner, François and Caccetta, Peter and Wu, Chen},
   year={2020},
   month=apr, pages={94--114}}

@inproceedings{hatamizadeh2021unetrtransformers3dmedical,
  title={{UNETR}: Transformers for 3-D medical image segmentation},
  author={Hatamizadeh, Ali and Tang, Yucheng and Nath, Vishwesh and Yang, Dong and Myronenko, Andriy and Landman, Bennett and Roth, Holger R and Xu, Daguang},
  booktitle={Proceedings of the IEEE/CVF winter conference on applications of computer vision},
  pages={574--584},
  year={2022}
}

@article{guillon2013geotime,
  title={Geotime: A 3-D automatic tool for chronostratigraphic seismic interpretation and filtering},
  author={Guillon, S{\'e}bastien and Keskes, Noomane and Gallon, Jonathan and Donias, Marc},
  journal={The leading edge},
  volume={32},
  number={2},
  pages={154--159},
  year={2013},
  publisher={Society of Exploration Geophysicists}
}

@INPROCEEDINGS{thouvenot2025geo25d,
  author={Thouvenot, Alexandre and Boillot, Lionel and Gripon, Vincent},
  booktitle={33rd European Signal Processing Conference (EUSIPCO)}, 
  title={Signal-Informed Semi-Supervised 3-D Segmentation for Subsurface Analysis}, 
  year={2025},
  volume={},
  number={},
  pages={621--625},
  doi={10.23919/EUSIPCO63237.2025.11226348}}

@article{yu2023faultSurvey,
    title = {Current state and future directions for deep learning based automatic seismic fault interpretation: A systematic review},
    journal = {Earth-Science Reviews},
    volume = {243},
    pages = {104509},
    year = {2023},
    issn = {0012-8252},
    doi = {https://doi.org/10.1016/j.earscirev.2023.104509},
    author = {Yu An and Haiwen Du and Siteng Ma and Yingjie Niu and Dairui Liu and Jing Wang and Yuhan Du and Conrad Childs and John Walsh and Ruihai Dong}
}
    
\end{document}